\documentclass[11pt,letterpaper]{article}
\usepackage{authblk} 
\usepackage{emnlp2017}
\usepackage{times}
\usepackage{url}
\usepackage{latexsym}
\usepackage[T1]{fontenc}    
\usepackage{hyperref}       
\usepackage{url}            
\usepackage{booktabs}       
\usepackage{amsfonts}       
\usepackage{nicefrac}       
\usepackage{microtype}      
\usepackage{amsmath}
\usepackage{graphicx,color,xcolor}
\usepackage{algorithm}
\usepackage[noend]{algpseudocode}
\usepackage{multirow}

\emnlpfinalcopy 

\newcommand{\imj}[1]{{\bfseries \small \color{red} #1}} 

\title{Neural Machine Translation with Gumbel-Greedy Decoding}
\def \nyu{$^\ddag$}
\def \hku{$^\dagger$}

 \author[\hku]{\bf Jiatao Gu}
 \author[\nyu]{\bf Daniel Jiwoong Im}
 \author[\hku]{\bf Victor O.K. Li}
 \affil[\hku]{The University of Hong Kong}
 \affil[\nyu]{AIFounded Inc.}
 \affil[\hku]{\tt  \{jiataogu, vli\}@eee.hku.hk}
 \affil[\nyu]{\tt  daniel.im@aifounded.com}

\date{}

\begin{document}
\maketitle
\begin{abstract}

Previous neural machine translation models used some heuristic search algorithms (e.g., beam search) in order to avoid solving the maximum a posteriori problem over translation sentences at test time. In this paper, we propose the \textit{Gumbel-Greedy Decoding} which trains a generative network to predict translation under a trained  model. We solve such a problem using the Gumbel-Softmax reparameterization, which makes our generative network differentiable and trainable through standard stochastic gradient methods. We empirically demonstrate that our proposed model is effective for generating sequences of discrete words. 

\end{abstract}
\section{Introduction}

Neural machine translation (NMT)~\cite{cho2014learning,sutskever2014sequence,bahdanau2014neural}, as a new territory of machine translation research, has recently become a method of choice, and is empirically shown to be superior over traditional translation systems.

The basic scenario of modeling neural machine translation is to model the conditional probability of the translation, in which we often train the model that either maximizes the log-likelihood for the ground-truth translation (teacher forcing) or translations with highest rewards (REINFORCE).
Despite these advances, a central problem that still remains with such sequential modeling approaches: once the model is trained, the most probable output which maximizes the log-likelihood during trained cannot be properly found at test time. This is because, it involves solving the maximum-a-posteriori (MAP) problem over all possible output sequences. To avoid this problem, heuristic search algorithms (e.g., greedy decoding, beam search) are used to approximate the optimal translation.

In this paper, we address this issue by employing a discriminator-generator framework -- we train the discriminator and the generator at training time, but emit translations with the generator at test time. Instead of relying on a non-optimal searching algorithm at test time, like greedy search, we propose to train the generator to predict the search directly. Such a way would typically suffer from non-differentiablity of generating discrete words. Here, we address this problem by turning the discrete output node into a differentiable node using the Gumbel-Softmax reparameterization~\citep{jang2016categorical}. Throughout the paper, we named this new process of generating sequence of words as {\em the Gumbel Greedy-Decoding} (GGD). 
We extensively evaluate the proposed GGD on a large parallel corpora with different variants of generators and discriminators. The empirical results demonstrate that GGD improves translation quality.

\section{Neural Machine Translation}
\label{sec.nmt}
Neural Machine Translation (NMT) models commonly share the auto-regressive property as it is the natural way to model sequential data. More formally, we can define the distribution over the translation sentence $Y = [y^1, ..., y^T]$ given a source sentence $X=[x^1, ..., x^{Ts}]$ as a conditional language model:
\begin{equation}
\label{eq.crnnlm}
p(Y|X) = \prod_{t=1}^T p(y^t|y^{<t}, X).
\end{equation}
The conditional probability is composed of an encoder $e_t(\cdot)$ and a decoder network $d_t(\cdot)$ with a softmax layer on top.  For notation, we denote the vocabulary size of the target language as $K$ and each  word $y^t$ is assigned to an index $k \in [1, K]$.  In this paper, we use the one-hot representation for each word, that is, $y^t_i=\mathbb{I}[i=k], i=1,...K$.
Thus the probability is computed using softmax:
\begin{equation}
\label{eq.model}
p(y^t|y^{<t}, X) = \text{softmax}\left[a\left(z^t; \theta_a\right)\right]^\top\cdot y^t 
\end{equation}
where $\text{softmax}(a)_i =\frac{\exp(a_i)}{\sum_{j=1}^K \exp(a_j)}$, and
\begin{equation}
z^t = f(z^{t-1}, y^{t-1}, e^t(X; \theta_e); \theta_d)
\end{equation}
$z_t$ is the hidden state of the decoder at step $t$, and $a$ is the energy function which maps the hidden state into a distribution over the vocabulary.
The output of the encoder $e^t(X)$ is a time-dependent feature of the source sentence $X$. Typically, both the encoder and decoder consist of deep recurrent neural networks (with the {\em soft} attention mechanism integrated) \cite{bahdanau2014neural,luong2015effective}. We use $\theta = \{\theta_a, \theta_d, \theta_e\}$ to denote the parameters of the NMT model.

\subsection{Training phase} 
There are two common ways to train NMT models, which are teacher forcing~\cite{williams1989learning,sutskever2014sequence,bahdanau2014neural} and REINFORCE~\cite{williams1992simple,ranzato2015sequence,shen2015minimum,bahdanau2016actor} algorithms~\footnote{For simplicity, previous efforts using reinforcement learning to train NMT are treated as variants of REINFORCE.}. 

In teacher forcing, the model is trained to maximize the conditional log-likelihood (MLE) of the ground-truth translation $Y^*$ given the source sentence $X$. In contrast, the REINFORCE algorithm does not rely on the ground-truth translation, but it maximizes the expectation of a global reward function $R$. 
In an unified view, the gradients w.r.t the parameters $\theta$ for both methods can be seen as:
\begin{equation}
\label{eq.mle}
\mathbb{E}_{Y\sim \mathcal{M}}\left[\frac{\partial}{\partial \theta}\log p_{\theta}(Y|X) \cdot R(Y)\right]
\end{equation}
where for teacher forcing, $\mathcal{M}$ is the empirical distribution on $Y|X$ and $R(Y)\equiv 1$, while for REINFORCE, $\mathcal{M}$ is $p_\theta$ itself and $R$ is used to re-weight the gradients.
The primary difference between teacher forcing and REINFORCE is that teacher forcing corrects the translation word-by-word based on the ground-truth prefix, whereas REINFORCE rewards the translated sentence as a whole. 
The training of teacher forcing is stable but it suffers from the local normalization property \cite{ranzato2015sequence}.
Whereas, although REINFORCE does not have such a problem, it is known to be difficult to train due to the high variance in its gradients.

\subsection{Test phase} 
At the test phase, our goal is to get the best translation of the source sentence possible.
This process is also known as the {\em decoding} process. Ideally,
we can use Maximum-a-Posteriori (MAP) to find a translation $Y$ which maximizes $\log p_\theta(Y|X)$.
Unfortunately, exact MAP inference is intractable due to the exponential complexity in searching. Therefore, we approximate the MAP inference based on some heuristic search-based methods in practice:

\paragraph{\textbf{Sampling \& Greedy Decoding}} 
As the model is learned, we can directly perform sampling from the conditional distribution word-by-word, in which case the translation is stochastic. In contrast, 
rather than maximizing the log-likelihood for the entire translation, greedy decoding simply picks the most likely word at each time step $t$, resulting in a deterministic translation. However, it is inadequate in practice due to lack of future information.  

\paragraph{\textbf{Beam Search}}  
Beam search usually finds better translation by storing $S$ hypotheses with the highest scores ($\prod_{t'=1}^t p(y^t | y^{<t}, X)$). When all the hypotheses terminate, it returns the hypothesis with the highest log-probability. 
Despite its superior performance compared to greedy decoding, the computational complexity grows linearly w.r.t. $|S|$, rendering it less preferable in production environment.

\section{Discriminator-Generator framework}
\label{sec.framework}
\begin{figure*}[htpb]
\vspace{-10pt}
\centering
\includegraphics[width=0.93\linewidth]{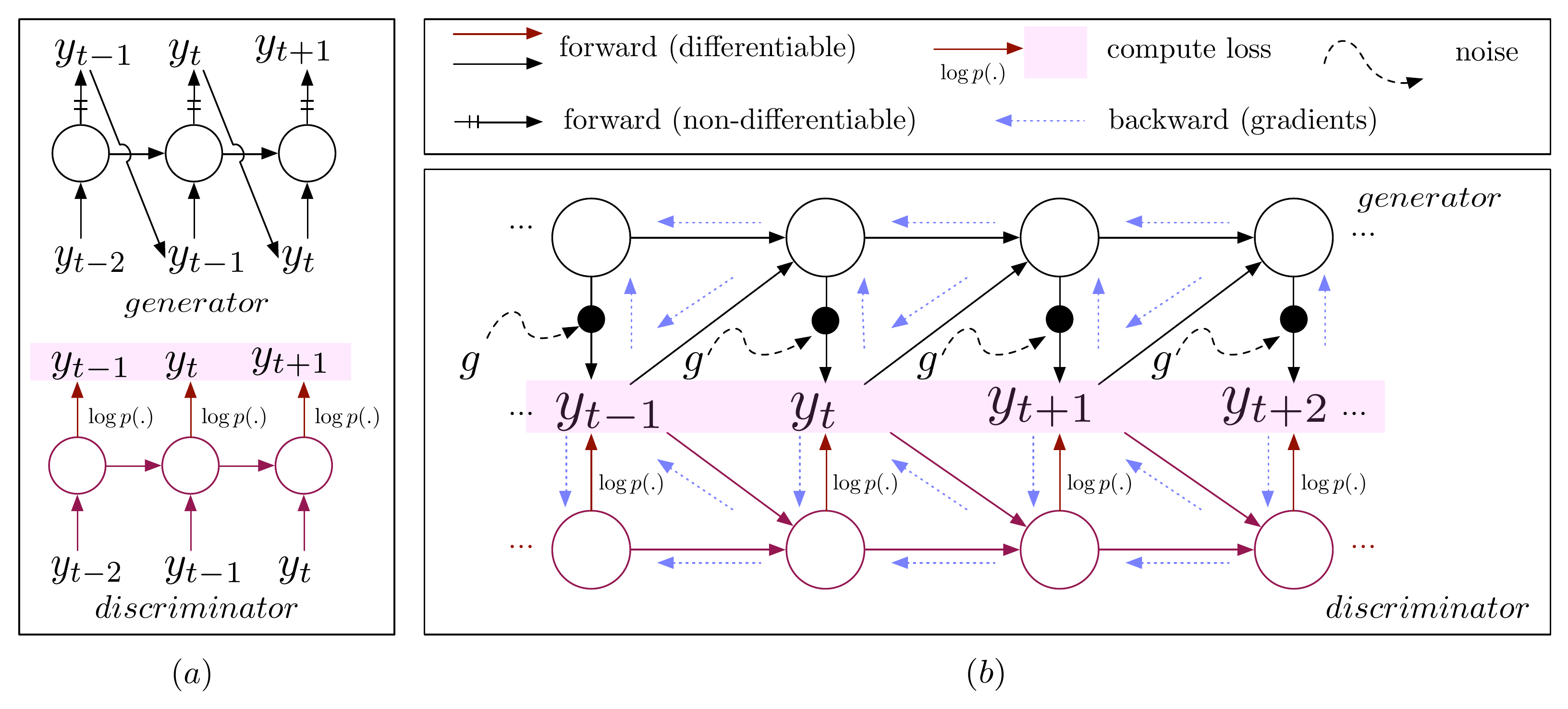}
\vspace{-15pt}
\caption{\label{fig.actor} \small (a) An example illustrating the two functions of a NMT model: \textit{discriminator}  and \textit{generator}, (b)  An illustration of computational flow for the training of \textit{Gumbel-Greedy Decoding}. The cost function uses the same architecture of the generator to compute the log-likelihood. }
\vspace{-15pt}
\end{figure*}

The major discrepancy between training and testing time is that we cannot leverage the full power of our trained NMT model during testing.
Here, we propose to train a separate generative network that will reduce the potential mismatch between the training and testing phases.
Let us first portray the training and test procedure in terms of the discriminator-generator framework as following: 
\begin{itemize}
\vspace{-3pt}
\item \textit{NMT-discriminator} - measures the log-likelihood at word level - $\log p_{\theta}(Y|X)$ - given the source sentence $X$ and a translation $Y$. 
\vspace{-5pt}
\item \textit{NMT-generator} - generates the translation by taking the output of the word as an input to next step recursively - $Y=G_{\phi}(X)$ - given the source sentence $X$. ($G$ is usually a search-based method).
\vspace{-7pt}
\end{itemize}
We train the generative network through a ``GAN-like'' discriminator-generator-framework, where the output of the NMT-generator\footnote{The term generator and decoder are used interchangeably.} gets fed to the NMT-discriminator (see Fig.~\ref{fig.actor}~(b)). We propose to learn the parameters of generator $\phi$ by maximizing the NMT discriminator's score,
\begin{equation}
\label{eq.framework}
J\left(\phi \right) = \mathbb{E}_{Y\sim G_\phi}\log p_\theta\left(Y|X\right)
\end{equation}
and the gradient w.r.t $\phi$ is computed using chain rule,
\begin{equation}
\label{eq.gradient}
\frac{\partial\log p_\theta\left(Y|X\right)}{\partial Y} \bigg|_{Y=G_\phi(X)} \cdot \frac{\partial G_\phi\left(X\right)}{\partial \phi}
\end{equation}

In practice, we can set the initial parameters of the generator to be the same as the discriminator's parameters.
Note that the discriminator and the generator share the same parameters, i.e., $\theta = \phi$, and the generator is never trained in the traditional NMT framework. 

Unfortunately, optimizing the generator with Eq.~\ref{eq.framework} and \ref{eq.gradient} involves operations, such as sampling or argmax, that are non-differentiable w.r.t (discrete) words. Therefore, we cannot leverage the backpropagation algorithm~\cite{rumelhart1986learning}. Here, we solve this problem by incorporating the Gumbel-Softmax relaxation into it(Section~\ref{sec:model}).

\section{Gumbel-Greedy Decoding}\label{sec:model}
In this section, we show how to train the generator w.r.t the discriminator's output using the idea of  {\em Gumbel-Greedy Decoding}, where we apply the Gumbel-Softmax
based reparameterization trick in sampling of the NMT-generator. The main idea is to turn the stochastic node (the last layer of the generator network) into a differentiable function of the network parameters with an independent random variable. 

\subsection{Sampling as Gumbel-Greedy Decoding}
\label{sec.sample}

\paragraph{The Gumbel-Max Trick}\newcite{gumbel1954statistical} transforms sampling from a categorical distribution to an optimization problem, and visa versa. That is to say, $y \sim p_\theta\left(y|y^{<t}, X\right) = \text{softmax}(a)$ in Eq.~\ref{eq.model} is equivalent to~\footnote{We omit the time-step mark $t$ for simplicity.}:
\begin{equation}
\vspace{-4pt}
y = \text{argmax} \left(g + a\right), g \sim \text{Gumbel i.i.d.}
\label{eq.argmax}
\end{equation}
where $\text{argmax}(x)_i= \mathbb{I}[x_i=\max(x)]$, and each element in $g$ can be computed using the inverse transform sampling of an auxiliary random uniform variable $u_i \sim \mathcal{U}(0,1)$, $g_i=-\log(-\log(u_i))$. 
Since the Gumbel noise $g$ and sampled words are independent, we can simply break down the sampling into a two-step process: 
\begin{enumerate}
\vspace{-7pt}
\item Sample a noise $g_t$ from Gumbel distribution at each time step;
\vspace{-7pt}
\item Perform the greedy decoding based on a noise-biased distribution in Eq.~\ref{eq.argmax}.
\vspace{-7pt}
\end{enumerate}
Note that the Gumbel-max trick does not change the non-differentiability of sampling,

\paragraph{Gumbel-Softmax Relaxation}
\newcite{maddison2016concrete,jang2016categorical} proposed a  reparameterization trick for discrete random variables based on Gumbel-Softmax where
\begin{equation}
\hat{y} = \text{softmax}((g + a)/\tau), g \sim \text{Gumbel i.i.d.}
\label{eq.softmax}
\end{equation}
where $\tau \in (0,\infty)$ is the temperature. The softmax function approaches argmax operations as $\tau \rightarrow 0$, and it becomes uniform when $\tau \rightarrow \infty$. Thus, the samples are no longer one-hot vectors. 
With the Gumbel-Softmax relaxation, we can easily derive the partial gradient estimator $\partial \hat{y}/\partial a$ of Gumbel-Softmax as:
\begin{equation}
	\frac{\partial \hat{y}_i}{\partial a_j} = \hat{y}_i \left(\delta_{ij}-\hat{y}_j\right)/\tau
    \label{eq.gummbel_grad}
\end{equation}
where $\delta_{ij}=\mathbb{I}[i=j]$. This allows us to train the NMT model using the backpropagation algorithm. Note that according to Eq.~\ref{eq.gummbel_grad}, $\lim_{\tau \rightarrow 0}\left[{\partial \hat{y}_i}/{\partial a_j}\right] = 0$ (or $\pm\infty$ if more than 2 words achieve the maximum energy simultaneously), which makes training with backpropagation impossible for $\tau \rightarrow 0$.

\paragraph{Straight-Through (ST) Gumbel}~
Nonetheless, there is still a remaining challenge to overcome before we can apply Gumbel-Softmax reparameterization to NMT. In language modeling, the embedding vector is chosen from the look up table based on the generated word, and is emitted in the next time step. However, the Gumbel-Softmax relaxation leads to a mixture of embedding vectors, in turn causing a mixing error. Furthermore, such mixing errors get accumulated over time as the errors are propagated forwards through the recurrent neural network. This causes future word generation to deteriorate even with a small temperature $\tau$, especially when we are using a pre-trained model. 

In order to avoid the problem of mixing and propagating word embedding errors over time, we apply {\em the straight-through} version of the Gumbel-Softmax estimator \cite{jang2016categorical}, or called ST-Gumbel. During the forward phase, we use the Gumbel-Max in Eq.~\ref{eq.argmax}, while computing the gradient of the Gumbel-Softmax in Eq.~\ref{eq.gummbel_grad}, i.e., $\hat{y}_t$ in Eq.~\ref{eq.gummbel_grad} is replaced by $y_t$ in Eq.~\ref{eq.argmax}. Obviously, the ST-Gumbel estimator is biased due to the sample mismatch between the forward and backward passes. However, we find that it helps, empirically. 

\paragraph{Learning} By putting all together, we can derive the basic learning algorithm for Gumbel-greedy decoding. We estimate the gradient from a differentiable cost function $R(Y)$ w.r.t.  $\phi$:
\begin{equation}
\label{eq.gumbel-learn}
\begin{aligned}
\mathbb{E}_{Y \sim G_\phi}\left[\frac{\partial}{\partial \phi} R(Y)\right] 
&\approx  \mathbb{E}_{g \sim P}\left[\frac{\partial R}{\partial Y} \frac{\partial \hat{Y}}{\partial \phi}\right]
\end{aligned}
\end{equation}
where we use $P$ to represent the noise distribution over all time steps. From the equation, it is clear that such approximation holds by assuming ${\partial Y}/{\partial \hat{Y}}\approx 1$ which however is not always true. In practice, we can use any differentiable cost function that evaluates the goodness of the generation output. For instance, a fixed language model, a critic that predicts BLEU scores, or the NMT-discriminator (Eq.~\ref{eq.framework}). We will discuss this in later sections.


\subsection{Arbitrary Decoding Algorithms as Gumbel-Greedy Decoding}
\label{sec.abs}
\paragraph{Inference on Gumbel} The Gumbel-Max trick indicates a general formulation that can present any  decoding algorithms as Gumbel-Greedy Decoding:
\begin{equation}
y = \text{argmax} \left(g + a\right), g \sim Q
\end{equation}
where $Q$ represents a special distribution that generates this word, which is typically unknown. Note that when we choose $Q=P$, the decoding algorithm degenerates into sampling. However, as discussed in~\newcite{maddison2014sampling}, given the trajectory of decoded words, we can efficiently infer its corresponding Gumbel noise $g$ using a top-down construction algorithm as $g = g^* - a$, and:
\begin{equation}
\label{eq.inference}
g^*_i =  
 \left\{
  \begin{aligned}
    &g',  
    \quad\quad\quad\quad\quad\quad\quad\quad
    y_i \text{is selected} \\
    &g'-\log\left[1 + e^{g' -\tilde{g}_i}\right], \text{otherwise} \\
  \end{aligned}
  \right.
\end{equation}
where $\tilde{g}_i = -\log(-\log(u_i)) + a_i$ and the ``top-gumbel" $g' = -\log(-\log(u)) + \log\left(\sum_i \exp(a_i)\right)$ for $ u, u_i\sim \mathcal{U}(0, 1)$, respectively. Such inference is also a special case of the core top-down algorithm of A* sampling~\cite{maddison2014sampling}.

\paragraph{Learning}
The above inference algorithm shows that, after running any decoding algorithm $G_{\phi}(X)$ (e.g. greedy decoding, beam search, etc.), we can always infer corresponding noise $g^t$ at each time step. Although in such cases the inferred noise does depend on the translation, which breaks the requirement of reparameterization trick, the decoding methods we use are usually deterministic methods. That is, $p\left(g,Y| X\right)\approx p\left(g|X\right)p\left(Y|X\right)$.  It is possible to train the deterministic generator as an equivalent Gumbel-greedy decoding using Eq.~\ref{eq.gumbel-learn}. 


\subsection{Gumbel-Greedy Decoding for Discriminator-Generator Framework}\label{subsec:GGD_DG}
We can finally conclude the learning algorithm for the proposed Discriminator-Generator framework using GGD, by simply setting the discriminator's output $\log p_\theta(Y|X)$ as the cost function in Eq.~\ref{eq.gumbel-learn}, as shown in Fig~\ref{fig.actor}(b), where we illustrate the computational flow of the whole framework. Note that the non-dfferentiable path is replaced by a differentiable path with an additional noise due to GGD, and gradient (though biased) flows can freely go through both directions of the discriminator and the generator, sufficiently communicating all useful information for learning. The overall algorithm for learning the generator using GGD is found in Algorithm 1.

\paragraph{With Regularization} One issue we observed in practice is that, directly optimizing the discriminator's output is not stable for learning the generator with GGD. Fortunately, we can stabilize the optimization by adding an entropy term in the cost function w.r.t $\phi$:
\begin{equation}
\label{eq.regularize}
 \mathbb{E}_{G_\phi}\left[\log p_\theta\left(Y|X\right)\right] - \mathbb{E}_{G_\phi}\left[\log p_{\phi'}\left(Y|X\right)\right]
\end{equation}
where we use $\phi'$ to represent a copy of the current parameters $\phi$ and make it as a ``discriminator". Note that gradients w.r.t $\phi$ will not flow into $\phi'$.

\paragraph{Adversarial Learning} Even though it is possible to learn the generator with a fixed discriminator, the proposed framework also allows to optimize both the discriminator and the generator in an adversarial way~\cite{goodfellow2014generative}. In particular, we take a similar formulation of the energy-based generative adversarial nets~\cite{zhao2016energy} where in our case we use the discriminator's output as the energy to distinguish the ground-truth translation and the generator's generation, w.r.t $\theta$:
\begin{equation}
\label{eq.gan}
\mathbb{E}_{D}\left[\log p_\theta\left(Y|X\right)\right] - \mathbb{E}_{G}\left[\log p_{\theta}\left(Y|X\right)\right]
\end{equation}
where $D$ is the empirical distribution of real translation. In practice, we alternate the training of the generator and the discriminator iteratively.

\begin{algorithm}[hptb]
\caption{Gumbel-Greedy Decoding}
\label{algo2}
\begin{algorithmic}[1]
\small
\Require{discriminator $p_{\theta}$, generator $G_{\phi}$, $N_d\geq 0$, $N_g>0$}
\State Train $\theta$ using MLE/REINFORCE on training set $D$;
\State Initialize $\phi$ using $\theta$;
\State Shuffle $D$ twice into $D_{\theta}$ and $D_{\phi}$
\While{stopping criterion is not met}
\For{$t=1:N_g$}     \quad\quad  // \textit{learn the generator}
\State Draw a translation pair: $(X, \_)\sim D_{\phi}$;
\State Obtain $Y, \hat{Y} = $ \textsc{GumbelDec}$(G, X)$
\State Compute forward pass $\sim X, Y$ with Eq.~\ref{eq.regularize}
\State Compute backward pass $\sim X, \hat{Y}$, update $\phi$ 
\EndFor
\For{$t=1:N_d$}     \quad\quad  // \textit{learn the discriminator}
\State Draw a translation pair: $(X, Y^*)\sim D_{\theta}$;
\State Obtain $Y, \_ = $ \textsc{GumbelDec}$(G, X)$
\State Compute forward pass $\sim X, Y, Y^*$ with Eq.~\ref{eq.gan}
\State Compute backward pass $\sim X, Y, Y^*$, update $\theta$ 
\EndFor
\EndWhile
\Statex{}
\setcounter{ALG@line}{0}
\Statex{\hspace{-18pt}\textbf{Function: }}{\textsc{GumbelDec}$(G, X)$}
  \If{$G = $ `sampling'}
    \State Sample $g \sim \text{Gumbel i.i.d.}$
	\State Obtain $Y, \hat{Y}$ with Eq.~\ref{eq.argmax} and Eq.~\ref{eq.softmax}
  \Else
  	\State Obtain $Y = G(X)$
    \State Infer $g$ with Eq.~\ref{eq.inference}
    \State Obtain $\hat{Y}$ with Eq.~\ref{eq.softmax}
  \EndIf
\State Return $Y, \hat{Y}$
\end{algorithmic}
\end{algorithm}

\section{Experiments}

We conduct the studies of learning to decode using the Gumbel-Greedy Decoding (GGD) technique as discussed in Section~\ref{sec:model}. We extensively compare our GGD-based neural machine translation decoding model with the traditional decoding methods such as greedy decoding and show its effectiveness. 

\subsection{Experimental Setup}
\label{sec.preliminary}
\paragraph{Dataset} We consider translating -- Czech-English (Cs-En) and German-English (De-En) -- language pairs for both directions with a standard attention-based neural machine translation system \citep{bahdanau2014neural}. We use the parallel corpora available from WMT'15\footnote{http://www.statmt.org/wmt15/} as a training set. We use newstest-2013 for the validation set to select the best model according to the BLEU scores and use newstest-2015 for the test set. All the datasets were tokenized and segmented into sub-word symbols using byte-pair encoding (BPE)~\citep{sennrich2015neural}. We use sentences of length up to 50 subword symbols for teacher forcing and 80 symbols for REINFORCE, GGD and testing.  

\paragraph{Architecture} We implement the NMT model as an attention-based neural machine translation model whose encoder and decoder recurrent networks have 1,028 gated recurrent units \citep[GRU,][]{cho2014learning} each. For the encoder, a bidirectional RNN is used and we concatenate the hidden states from both directions to build the context at each step. For the decoder, a single layer feed-forward neural network (512 hidden units) is used to compute the attention scores. Both source and target symbols are projected into 512-dimensional embedding vectors. The same architecture is shared by the NMT-discriminator and the NMT-generator. 

\paragraph{Baselines}  
We set our baseline to be NMT model trained with teacher forcing and REINFORCE algorithm. 
Our NMT model was trained with teacher forcing method (Maximum Likelihood) for approximately 300,000 updates for each language pairs. These networks were trained using Adadelta~\citep{zeiler2012adadelta}. We further fine-tuned these models using REINFORCE with a smoothed sentence-level BLEU~\citep{lin2004automatic} as reward following similar procedures in ~\newcite{ranzato2015sequence}.  We denote the former trained model as $\theta_{ML}$ and the additionally trained model using REINFORCE as $\theta_{RL}$. 

Additionally, we explored the Straight-Through (ST) estimator~\cite{bengio2013estimating,chung2016hierarchical} and compared with the ST-Gumbel that GGD uses for passing the gradients. The difference between the two is that, we use the output distribution $\text{softmax}\left(a/\tau\right)$ instead of $\text{softmax}((g + a)/\tau)$ in Eq.~\ref{eq.softmax} in the original ST estimator. The ST estimator, as a special case of ST Gumbel estimator, is independent of the choice of the selected word in the forward pass.

\paragraph{Pre-training} In our experiments, we use pre-trained models from the baseline $\theta_{ML}$ for the discriminative networks for training generative (decoding) network.
It is possible to start a generator $\phi$ from scratch for generating translation, and yet it has been shown to be effective if the generator is continually learned from the initialization of a pre-trained discriminator~\cite{ranzato2015sequence,shen2015minimum,bahdanau2016actor,lamb2016professor}. Because our learning algorithm requires sampling from the generator, the searching space is extensive for a randomly initialized generator to output any meaningful translation to learn from. In our experiments, we observed that initializing the parameter of the generator $\phi=\theta_{ML}$ worked consistently better whether we choose a stochastic generator for sampling or a deterministic one for greedy decoding. 

\paragraph{Learning of GGD}
We report the results of generator trained with sampling and greedy decoding, respectively based on Eq.~\ref{eq.regularize}. We find that learning using RMSProp~\citep{tieleman2012lecture} is most effective with the initial learning rates of $1\times 10^{-5}$. It is also possible to continually learn the discriminator according to Eq.~\ref{eq.gan}. The generator usually gets updated much more than the discriminator. In our experiments, we used 10 updates for the generator for every discriminator's update.

We denote GGD-GAN for the model where the discriminator and the generator is jointly trained. We denote GGD-Fixed-D
for the model where only the generator is trained with fixed discriminator. 

\begin{figure}[t]
\includegraphics[width=\linewidth]{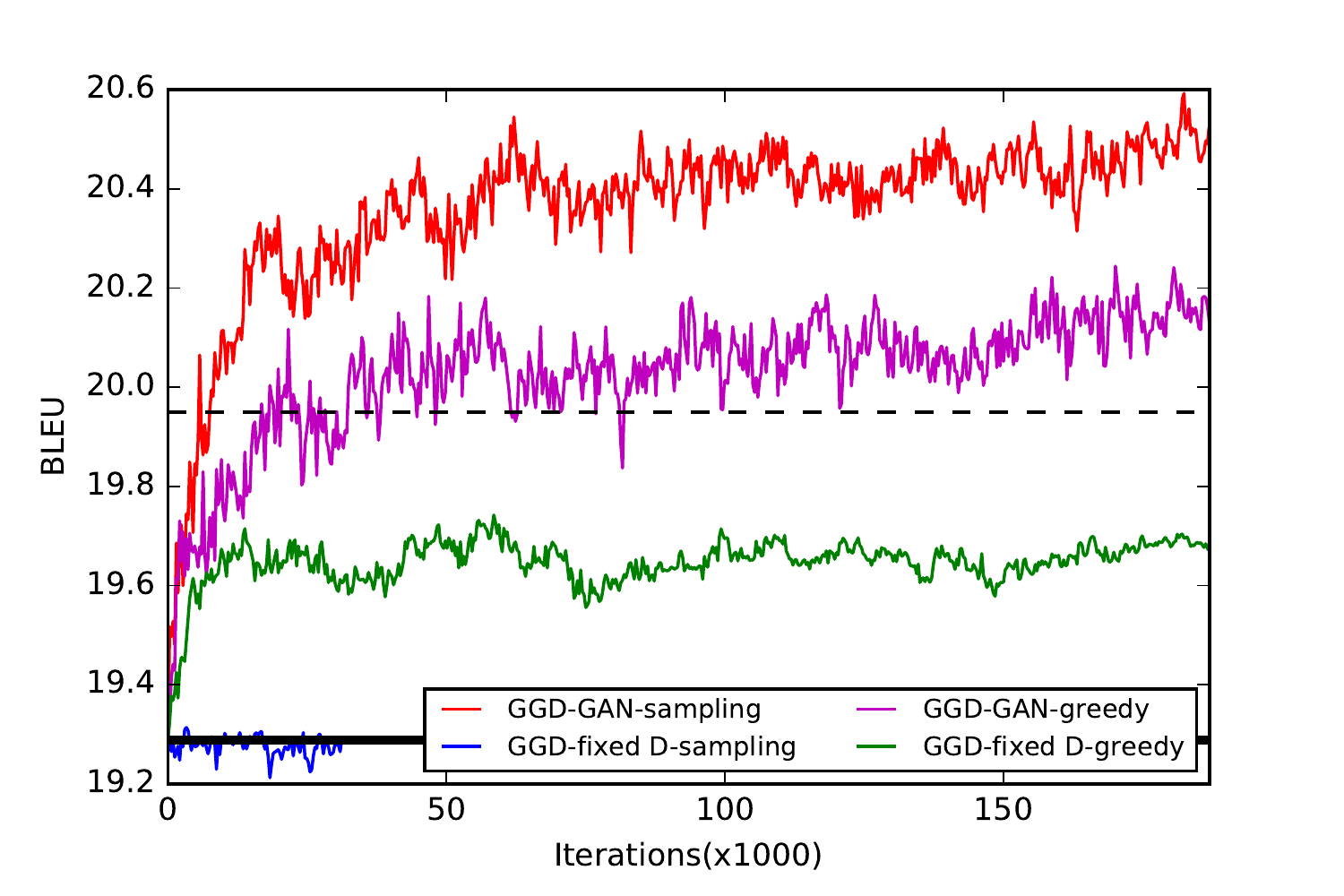}
\vspace{-7mm}
\caption{\label{fig.kt1}\small Comparison of greedy BLEU scores on the validation set of Cz-En, achieved by two generators that are learned to optimize a discriminator initially trained with teacher forcing. ``GAN'' refers to the discriminator being iteratively trained together with the generator, while  ``fixed D'' refers to the discriminator being fixed.
The straight black line and the black dashed lines are the BLEU scores achieved by the baseline models learned with teacher forcing and REINFORCE.} 
\vspace{-3mm}
\end{figure}


\subsection{Results and Analysis}
\begin{figure*}[htb]
\includegraphics[width=\linewidth]{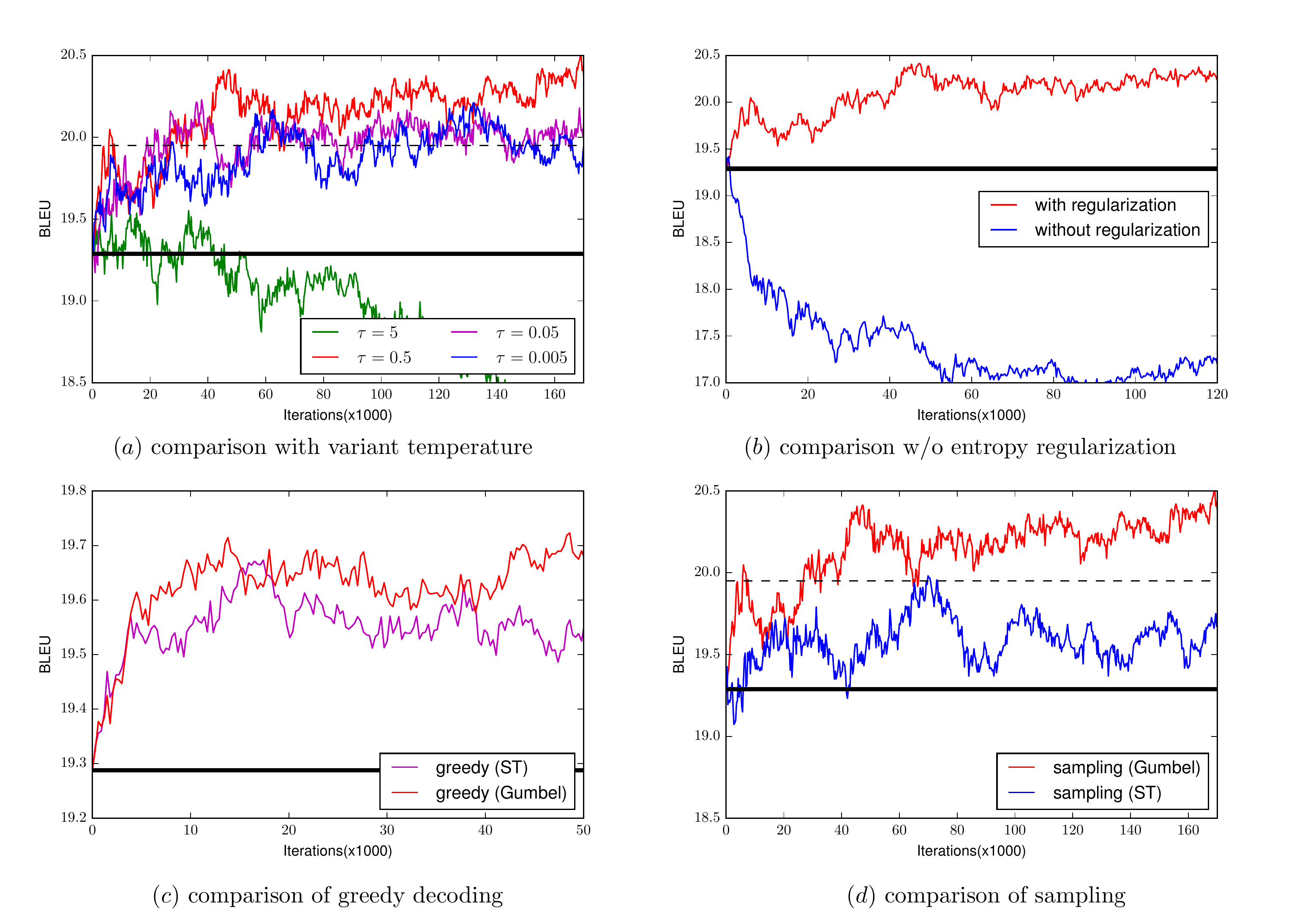}
\vspace{-7.5mm}
\caption{\label{fig.kt} \small Comparison of greedy BLEU scores on the validation set of Cs-En. Both (a) and (b) are achieved by stochastic generators that are learned to optimize a discriminator trained with REINFORCE. (c) shows the comparison of learning the generator of greedy decoding w.r.t. the teacher-forcing discriminator; (d) shows the comparison of learning the generator of sampling w.r.t. the REINFORCE discriminator. For all sub-figures, the black straight line and the black dashed line mean the BLEU scores achieved by the baseline models learned with teacher forcing and REINFORCE, respectively.} 
\end{figure*}

In our first experiment, we examine whether the GGD-GAN is more effective compare to GGD-fixed D.
Fig.~\ref{fig.kt1} presents the results of training based on both sampling and greedy methods.  We observe that both GGD-GAN and GGD-Fixed-D give much higher than the two baseline models, $\theta_{ML}$ and $\theta_{RL}$, by $\approx+1.3$ and $\approx +0.6$ respectively. 
Furthermore, the curves in Fig~\ref{fig.kt1} shows that we get the best validation BLEU score when the discriminator is trained together with a stochastic generator with a adversarial loss.
The reason why GAN style of training works better than the fixed discriminator training is because, we cannot get any additional information that helps the generator when we are just training the generator (see the blue curve in Fig.~\ref{fig.kt1}) when we start with the same generator and the discriminator. 

Importantly, we notice that the generator with GGD always improves the score compared to the original model. This illustrates that even when the trained discriminator is not optimal, the discriminator can be jointly trained with the generator again to achieve better score. In fact, just by training greedy decoding on generator enhances the BLEU score as shown in Fig.~\ref{fig.kt} green curve.
Finally, we get the most improvement when we use GGD with sampling instead of GGD with greedy decoding.

\paragraph{Importance of Regularization}
We experimentally demonstrate the effectiveness of entropy regularization  from Section~\ref{subsec:GGD_DG}. Fig.~\ref{fig.kt} (b) presents the performance with and without the entropy regularization term. This figure illustrates that the generator drops dramatically and only optimize the discriminator when the entropy term is removed. 
We hypothesize that one reason could be that the output distribution of a pre-trained generator became highly peaked, and therefore, it is sensitive to small changes. 
Thus, just relying on a discriminator network, which act as a positive force that {\em pushes} the distribution go to a better direction is not sufficient. Rather, we need the regularizer that act as a negative force, which distributes the probability mass, is necessary.
Lastly, we also note that Eq.~\ref{eq.regularize} can also be seen as the minimizing the Kullback-Leibler (KL) divergence between $p_{\phi}$ and $p_{\theta_{RL}}$ and it achieves the optimal as $p_{\phi} = p_{\theta_{RL}}$.

\paragraph{The sensitivity analysis w.r.t the temperature $\tau$} 
One of the extra hyperparameter that is added from GGD is the Gumbel-Softmax temperature $\tau$ rate. Here, we explore how the changes in the temperature effect the performance. The four different temperature rates $\{5, 0.5, 0.05, 0.005\}$ were used in the experiment.

The curves in Fig.~\ref{fig.kt} (a) demonstrate that the best result is achieved when $\tau=0.5$. A smaller $\tau$ leads to a vanishing gradient problem. In contrast, a larger $\tau$ also leads to unstable training. We see that the performance curve drops dramatically at $\tau= 5$. We speculate that this is due to the bias in the estimator. As the bias inside the estimator depends on the amount of forward-backward mismatch $\Delta y=y-\hat{y}$, which is  proportional to the temperature we use.
\citet{jang2016categorical} suggests to anneal the temperature over the training time. However, we did not find the annealing technique help in practice at least for NMT.
All of our models were trained with temperature rate of $0.5$ in the other experiments.

\paragraph{ST versus ST-Gumbel} Last but not least, we compared the original Straight-Through (ST)
estimator~\cite{bengio2013estimating,chung2016hierarchical} with ST-Gumbel. Since ST is just a special case of ST-Gumbel, we can run all the experiments in the same way and simply drop the Gumbel noise term when computing the backward pass. As shown in Fig.~\ref{fig.kt} (c) and (d), we have two experiments using these two estimators, training greedy decoding and sampling, respectively. 

We observe that ST-Gumbel works better than the original ST estimator in both cases, especially when training the generator with sampling. This is because the backward pass of the ST-estimator is independent of the word that we choose in the forward pass. This is especially problematic for sampling-based compare to greedy-based, because we can get a sampled word that has a relatively small probability in the output distribution. In contrast, the ST-Gumbel always sets the selected word with the highest score in the Gumbel-Softmax (Eq.~\ref{eq.softmax}) by adding the noise. Consequently, this reduces the bias compared with the ST-estimator and makes the learning more stable.

\begin{table}
  \centering
  {\small
  \begin{tabular}{l|ccccc}
	
      Model  & DE-EN & EN-DE & CS-EN & EN-CS \\
      \hline\hline
        MLE          & 21.63   & 18.97 & 18.90  & 14.49   \\
        REINFORCE    & 22.56	 & 19.32 & 19.45  & 15.02     \\
        GGD-GAN      & 22.47   & 19.38 & 20.12  & 15.4\\ \hline
  \end{tabular}
  \caption{\small The greedy decoding performance of models trained with GGD-GAN against MLE and REINFORCE. BLEU scores are calculated on the test sets for all the language pairs.}
  \label{tab:GGD-greedy}
  }
\end{table}
 
\begin{table}
 \centering
 {\small
  \begin{tabular}{l|ccccc}
      Model  & DE-EN & EN-DE & CS-EN & EN-CS \\\hline\hline
        MLE                  & 24.46	& 21.33	& 21.2	& 16.2  \\
        GGD-GAN         	 & 25.32	& 21.27	& 21.17	& 16.44\\ \hline
  \end{tabular}
    \caption{\small The beam-search performance of models trained with GGD-GAN against MLE. BLEU scores are calculated on the test sets for all the language pairs.}
  \label{tab:GGD-beam}
  }
  \vspace{-10pt}
\end{table}

\paragraph{Final Results} 
Based on the above experiments, we find the most proper training setting for GGD is when we i) jointly training the discriminator and the generator, and ii) use sampling-based generator with additional entropy regularization. We report the final performance on all four language pairs with these settings in Table~\ref{tab:GGD-greedy} and \ref{tab:GGD-beam}/ 
Both BLEU scores of greedy decoding and beam-search (size=5) are reported. It is clear that the generators trained with the proposed GGD algorithm can consistently outperforms the baseline models and generates better translations.

\section{Related Work}
There has been several work on training to solve decoding problem in NLP \cite{shen2015minimum,ranzato2015sequence,wiseman2016sequence}. 
Recently, there has been a work that came out independently of ours on learning to decode.
\citet{li2017learning} proposed to train a neural network that predicts an arbitrary decoding objective given a source sentence and a partial hypothesis or a prefix of translation. They use it as an auxiliary score in beam search. For training such a network, referred to as a Q network in their paper, they generate each training example by either running beam search or using a ground-truth translation (when appropriate) for each source sentence. This approach allows one to use an arbitrary decoding objective, and yet it still relies heavily on the log-probability of the underlying neural translation system in actual decoding.

The proposed framework and the GGD algorithm are also directly motivated by 
Generative Adversarial Networks (GANs), which are one of the popular generative models that consist of discriminative and generative networks \cite{goodfellow2014generative}. Energy-based GAN was later introduced, which uses the energy as the score function rather than binary score (i.e., predicting whether the input is real or fake) \cite{zhao2016energy}.
The GAN style of training has been widely applied in vision domain~\cite{radford2015unsupervised,im2016generating, im2016gap}. There are only few works where GAN style of training is applied to sequential modeling \cite{Yu2016SeqGAN,2016arXiv161104051K} and machine translation. 
To the best of our knowledge, we are the first to apply Gumbel-softmax relaxation in a generator-discriminator framework for training neural machine translation. 
The closest work to ours is \newcite{2016arXiv161104051K}, which applies GAN for modeling simple sequences, and they also applied the Gumbel-Softmax to GAN. However, their problem setup and the training framework differ from ours in a sense that, i) their discriminator is exactly the same as the classical GAN, whereas our GAN is more close to the energy-based GAN; ii) they only apply to synthetic dataset, whereas we apply it to NMT with a large scale parallel corpora. The application of Gumbel distribution can also be seen in ~\cite{papandreou2011perturb}.

\section{Conclusion}
We studied way to learn a separate decoder for translating languages at test time. Our solution was to use the Gumbel-Softmax reparameterization trick, which makes our generative network differentiable and can be trained through standard stochastic gradient methods. We empirically demonstrate that our proposed model is effective for generating sequence of discrete words.

In the future work, we hope to explore adversarial learning using different reward functions with GGD. This includes both differentiable and non-differentiable rewards. As well, we can explore training the GAN network from scratch. 

\bibliography{acl2017}
\bibliographystyle{emnlp_natbib}

\end{document}